\documentclass[conference]{IEEEtran}
\IEEEoverridecommandlockouts
\usepackage[table,xcdraw]{xcolor}
\usepackage{tabularx,booktabs}
\usepackage{hyperref}
\usepackage{cite}
\usepackage{amsmath,amssymb,amsfonts}
\usepackage{algorithmic}
\usepackage{graphicx}
\usepackage{makecell}
\usepackage{textcomp}
\usepackage{xcolor}
\usepackage{float}
\usepackage{etoolbox}
\usepackage{atbegshi}
\AtBeginDocument{\AtBeginShipoutNext{\AtBeginShipoutDiscard}}
\makeatletter
\patchcmd{\@makecaption}
  {\scshape}
  {}
  {}
  {}
\makeatother
\def\BibTeX{{\rm B\kern-.05em{\sc i\kern-.025em b}\kern-.08em
    T\kern-.1667em\lower.7ex\hbox{E}\kern-.125emX}}
\begin{document}

\title{Explainable Deep Learning to Profile Mitochondrial Disease Using High Dimensional Protein Expression Data\\
{\footnotesize }
\thanks{EPSRC Centre for Doctoral Training in Big Data for Cloud Computing, Newcastle University}
}\textbf{}

\makeatletter
    \newcommand{\linebreakand}{%
      \end{@IEEEauthorhalign}
      \hfill\mbox{}\par
      \mbox{}\hfill\begin{@IEEEauthorhalign}
    }
    \makeatother

\author{\IEEEauthorblockN{ Atif Khan}
\IEEEauthorblockA{\textit{School of Computing, WCMR} \\
\textit{Newcastle University}\\
Newcastle upon Tyne, UK \\
a.khan21@newcastle.ac.uk}
\and
\IEEEauthorblockN{Conor Lawless}
\IEEEauthorblockA{\textit{WCMR} \\
\textit{Newcastle University}\\
Newcastle upon Tyne, UK \\
conor.lawless@newcastle.ac.uk}
\and
\IEEEauthorblockN{Amy E Vincent}
\IEEEauthorblockA{\textit{WCMR} \\
\textit{Newcastle University}\\
Newcastle upon Tyne, UK \\
amy.vincent@newcastle.ac.uk}
\and
\linebreakand
\IEEEauthorblockN{Satish Pilla}
\IEEEauthorblockA{\textit{School of Computing} \\
\textit{Newcastle University}\\
Newcastle upon Tyne, UK \\
s.pilla2@newcastle.ac.uk}
\and
\IEEEauthorblockN{Sushanth Ramesh}
\IEEEauthorblockA{\textit{School of Computing} \\
\textit{Newcastle University}\\
Newcastle upon Tyne, UK \\
s.shivpura-ramesh2@newcastle.ac.uk}
\and
\IEEEauthorblockN{A. Stephen McGough}
\IEEEauthorblockA{\textit{School of Computing} \\
\textit{Newcastle University}\\
Newcastle upon Tyne, UK \\
stephen.mcgough@newcastle.ac.uk}
}

\maketitle

\begin{abstract}
Mitochondrial diseases are currently untreatable due to our limited understanding of their pathology. We study the expression of various mitochondrial proteins in skeletal myofibres (SM) in order to discover processes involved in mitochondrial pathology using Imaging Mass Cytometry (IMC). IMC produces high dimensional multichannel pseudo-images representing spatial variation in the expression of a panel of proteins within a tissue, including subcellular variation. Statistical analysis of these images requires  semi-automated annotation of thousands of SMs in IMC images of patient muscle biopsies. In this paper we investigate the use of deep learning (DL) on raw IMC data to analyse it without any manual pre-processing steps, statistical summaries or statistical models. For this we first train state-of-art computer vision DL models on all available image channels, both combined and individually. We observed better than expected accuracy for many of these models. We then apply state-of-the-art explainable techniques relevant to computer vision DL to find the basis of the predictions of these models. Some of the resulting visual explainable maps highlight features in the images that appear consistent with the latest hypotheses about mitochondrial disease progression within myofibres.

\end{abstract}


\section{Introduction}
Mitochondrial diseases are individually uncommon but are collectively the most common metabolic disorder affecting 1 in 5,000 people\cite{Warren2020DecodingCytometry}. They can cause severe disabilities and adversely affect the life expectancy of patients \cite{Barends2016CausesDisease}. They manifest either as a result of mutations in genes encoded by the mitochondrial DNA (mtDNA) and/or in genes encoded by the nuclear DNA (nDNA) whose products are imported into mitochondria \cite{Warren2020DecodingCytometry}. Mitochondrial disease pathology is complex and highly heterogeneous. Diagnosis usually requires algorithmic analysis of clinical history and of results from multiple laboratory investigations \cite{Barends2016CausesDisease}. Some of the latest approaches to classify mitochondrial diseases and quantify disease severity are based on the analysis of single cell protein expression data and clinical information from patients \cite{Alston2017TheDisease}.

The Wellcome Centre for Mitochondrial Research (WCMR) is one of the leading institutes conducting research into mitochondrial diseases worldwide, and we have an unparalleled repository of clinical data and tissue from controls and patients with mitochondrial disease \cite{WellcomeUK}. The data includes images of tissue sections that capture spatial variation in protein expression within tissue (including within cells) and from which single-cell average level protein expression can be measured. We use advanced protein expression measurement techniques like Image Mass Cytometry (IMC) that allow us to observe the spatial variation in the expression of up to 40 proteins in tissue simultaneously. Established statistical approaches to analyse this multiplexed high dimensional data can successfully identify defective myofibres and proportion of defective fibres seems to usefully segreagate patients from control subjects.  However, these approaches are based around statistical summaries of intensity per cell and fairly crude quantitative measures of cell morphology.  In this work, in order to derive as much information as possible from these rare and valuable patient data, we want to use DL analysis of raw images to learn more from protein expression pseudo-images.

Deep learning (DL) in medicine and healthcare is a rapidly advancing domain, making groundbreaking strides in diagnosis, prognosis and drug discovery \cite{2019AscentMedicine}. This is particularly noticeable in bioimaging where DL is facilitating a paradigm shift in detecting disease \cite{Meijering2020AAnalysis}. The adaptation of general computer vision DL models such as Convolutional Neural network (CNN), R-CNN (region based CNN) and biomedical specific DL models such as UNet are driving the invention of novel and groundbreaking DL pipelines for diagnosis and prognosis \cite{AnTechnology}. The use of DL to discover underlying natural phenomena is at the cutting-edge of AI research \cite{RoyalSocietyTHERESEARCH}\cite{Zobeiry2019ANPREPRINT} but we could not find any literature describing such applications in medicine i.e. the use of DL to discover clinical phenomena such as disease pathology.

To profile protein expression patterns associated with various mitochondrial diseases using DL is doubly challenging. First, we need to build models that can accurately predict different classes of mitochondrial disease. Second, we need to interrogate those models to understand the basis for those predictions. In this paper we explore relevant DL models to predict one relevant mitochondrial disease tissue class (i.e. does the subject have a mitochondrial disease) and more importantly explore available explainable techniques that can discover protein expression patterns associated with general mitochondrial disease.
The main contributions of this paper are as follows.

\begin{itemize}
\item Selecting and training an accurate DL model for our multiplexed, high-dimensional protein expression data.
\item Implementing a range of explainable techniques to find predictive protein expression patterns.
\item This work has been carried out by an interdisciplinary team of scientists that includes biomedical experts in mitochondrial disease. We provide a holistic assessment of the utility of these techniques in discovering underlying mitochondrial disease pathology.
\end{itemize}
 
\section{Background and Related Work}

\subsection{Background}

Mitochondria are organelles that produce ~90\% of the energy consumed within each of the trillions of cells that make up a human body \cite{Bianconi2013AnBody}. Dysfunction in mitochondria disproportionately affects cells with a high energy demand e.g., muscle cells and neurons. Mitochondria are unusual in that they have their own DNA (mtDNA). Genes in mtDNA code exclusively for mitochondrial proteins, but most mitochondrial proteins are encoded in nuclear (nDNA). Mutations affecting mitochondrial proteins, whether coded in mtDNA or nDNA manifest as mitochondrial diseases \cite{Alston2017TheDisease}. In the clinic, mitochondrial diseases are classified based on their genetic aetiology, i.e., the source and location of their mutation, as inherited or sporadic nDNA and/or mtDNA diseases.

mtDNA diseases progress independently in individual cells via the accumulation of genetic mutations in mitochondrially encoded genes (mtDNA mutations). When mtDNA mutations reach high proportions in an individual cell, this results in alterations in the proportions of mitochondrial proteins and associated subunits of the mitochondrial respiratory chain (RC) complexes which in turn results in mitochondrial dysfunction \cite{Lawless2020TheMutations}.

At the WCMR, images of skeletal muscle tissue sections, representing the expression of RC proteins, are captured using IMC. IMC is a recently developed method allowing quantitative analysis of protein levels in a highly multiplexed way, at single cell and subcellular resolution. IMC data consists of a stack of 16 bit pseudo-images with a resolution of 1$\mu$m x 1$\mu$m per pixel saved in OME-TIFF(.ome.tiff) format \cite{Corporation2019MCDFLDM-400317}.  Each of up to 40 images in the stack corresponds to the expression of a unique protein.

For downstream analysis of skeletal muscle tissue sections, currently individual SMs are identified using semi-automatic image segmentation. This step is carried out using a python image analysis package called mitocyto built around OpenCV \cite{CnrLwlss/mitocyto:Sections} which also computes statistical summaries of protein expression as well as morphological summaries (e.g. mean pixel intensity, area, aspect ratio, circularity). Average protein expression levels in single SMs are analysed using relevant statistical techniques\cite{CnrLwlss/mitocyto:Sections}\cite{PlotIMCData}. Comparisons are drawn between average protein expression levels in single SMs from matched patient \& control groups using relevant statistical models (e.g. linear regression, Gaussian Mixture Models (GMM))
\cite{CnrLwlss/mitocyto:Sections}\cite{PlotIMCData}\cite{Ahmed2022QuantifyingM.3243AgtG}

These existing techniques are powerful but in this paper we take a different approach, applying DL methods directly on raw, unsegmented protein expression image data with an aim to discover novel protein expression patterns.  For example, we expect that we might find tissue level, subcellular level or cell morphology patterns that our current methods necessarily ignore. We want to investigate if this is an appropriate challenge for DL models given their reported superior performance over other methods for high dimensional computer vision tasks \cite{georgiou2020survey}.

\subsection{Deep Learning Models and Transfer Learning for Image Classification}
DL models based on CNN architecture have achieved great success in image classification tasks in various domains \cite{GirshickRichv5}\cite{krizhevsky2017imagenet}\cite{Li2022DeepSurvey}\cite{Masi2019DeepSurvey}\cite{Mazurowski2019DeepMRI}. Development of these models has been popular in the last decade leading to the invention of many new models that exponentially improved performance, setting new records for prediction accuracy on large public image datasets like imagenet \cite{PlestedDeepSurvey}. These models typically use conv+pool layers with a large number of filters or other layers or use techniques such as dropout, batch normalisation, ReLu activation, inception modules and residual learning to alleviate problems such as overfitting and vanishing gradient. These initial layers are typically followed by dense fully connected layers and an output layer that uses an activation function (mostly a softmax) to convert logit into output probabilities \cite{Li2021AProspects}. One way of tracking best models over the years has been to track the winners of imagenet’s ILSVRC challenge which highlighted models such as AlexNet, ZFNet, VGG, GoogLeNet and ResNet that are widely used in image classification applications today \cite{Russakovsky2015ImageNetChallenge}.
While the architectures of these DL models are the main cause of their improvements, these models need massive training data e.g. 14 million images in the imagenet dataset to achieve these results \cite{Li2021AProspects}. There are many real world cases (including ours) where availability of such data is impractical.
 In such cases transfer learning can usually provide better results than training with only a small dataset. Transfer learning is a method where the model is first trained on a related large dataset and the trained weights are used as a preset to further train the model with the original small dataset. It is usually prescribed that the related large dataset has closeness to the original data for the model to perform correctly\cite{Li2021AProspects}.

\subsection{Overview of Explainability of DL Models}

In machine learning and artificial intelligence, the terms explainability and interpretability are frequently used interchangeably. Despite how similar they appear, it is crucial to recognise the distinctions. Rudin \cite{RudinStopInstead} distinguishes between interpretable and explainable AI: while explainable AI attempts to offer post-hoc explanations for currently used black-box models, which are incomprehensible to humans, interpretable AI focuses on building models that are intrinsically explainable. Lipton \cite{LiptonTheInterpretability} emphasises the distinction between the questions that each family of techniques seeks to answer, interpretability asks, “How does the model work?” while explanation methods seek to respond, “What else can the model tell me?”. There is no general agreement on what either interpretability or explainability means \cite{Marcinkevics2020InterpretabilityMini-tour}, however, we have used Rudin’s definitions of these terms in this paper.

\subsection{Explainable Methods}
These methods, which are visualisations of the spatial variation in importance of input images with respect to prediction, provide post-hoc explanations. These methods can be categorised into three types namely function, signal and attribution visualisation. These groups of methods present different information about model prediction that are complementary to each other\cite{Kindermans2019TheMethods}. 

\paragraph {Function methods- [Gradients,Class Activation Map(CAM)]}: These methods explain using sensitivity analysis. Guided by model gradients these methods estimate how moving in a particular direction in the input space affects the output. To simplify, for a linear model y = w*x, these methods reduce to analyse weights(w) as gradient between output(y)  and input (x), $\partial$y/$\partial$x=w \cite{Kindermans2019TheMethods}\cite{Kindermans2019TheMethods}.  
Function methods like CAM were used to detect acute intracerebral haemorrhage, Lee et.al \cite{Lee2019AnDatasets} constructed a CAM from the output of an ensemble of CNNs: VGG-16, Inception-V3 and ResNet-50.

\paragraph{Signal methods- [Deconvnet, Guided Backprop]} These methods aim to isolate input patterns that simulate neuron activity in the higher layers of CNN i.e. analysing the components of the input data that causes the output \cite{Kindermans2019TheMethods}\cite{Alber2019INNvestigateNetworks}. Again to understand this consider a linear model y=w*x, for it the signal s = a * y where a = pattern that contains signal direction, intuitively it tells us where a change of output variable is expected to be measurable in the input \cite{Li2021AProspects}. Signals are more informative than functions in that they tell us both the regions and direction of the input image that are used by the model to predict output\cite{Li2021AProspects}. 
\\Signal methods like Deconvnet, Guided Backprop and Guided-Grad-CAM are used in the analysis of medical imaging data. For example, \cite{DeVos2019DirectCT} assessed coronary artery calcium for each slice of a heart or chest computed tomography (CT) image and applied deconvolution to reveal where in the slice the decision was made. Ji et.al \cite{Ji2019Gradient-basedImages} and Kowasri et.al \cite{Kowsari2020HMIC:Approach} employed Grad-CAM and demonstrated the use of a classifier to determine metastatic tissue in histological lymph node sections and identify small bowel enteropathy on histology respectively.

\paragraph{Attribution methods-[Deep Taylor, Input Gradient, Layer-wise Relevance propagation(LRP-Epsilon, Z, PresetAFlat, PresetBFlat)]}
These methods attribute importance to input features/signal dimensions for the output i.e. how much the signal dimensions/features of the input contribute to the output across the neural network \cite{Alber2019INNvestigateNetworks}\cite{Kindermans2019TheMethods}. For a linear model y =w*x attribution  r = ([w] $\otimes$ a) where [w] is weight vector,  $\otimes$ denote element-wise multiplication and a is signal \cite{Kindermans2019TheMethods}. Attributions are built upon signals i.e. attribution tells us the importance of each signal dimension/feature of the input image toward predicting the output, this is sometimes also referred to as relevance \cite{Kindermans2018LearningPatternattribution}.  
Attributions give more detailed explanation about model prediction than signal and these methods are used to analyse many medical DL models\cite{Bohle2019Layer-wiseClassification}. Bohle et.al \cite{Bohle2019Layer-wiseClassification} employed LRP (an attribution method) to locate Alzheimer’s disease-causing areas in brain MRI images. They contrasted the saliency maps produced by guided backpropagation with LRP and discovered that LRP was more accurate in detecting areas known to have Alzheimer’s disease.

All the above methods will provide perfect explanations for linear models but DL models are highly non-linear, which means these explanations can only be used as approximations \cite{Kindermans2019TheMethods}. To come back to our point that these methods complement each other to find explanations about model prediction, function methods help extract signal dimensions(features) from the data and attributions tell us the importance of each of these signal dimensions (features) toward predicting the output.

\begin{table*}[ht]
    \centering
    \begin{tabular}[width=1\textwidth]{|c|ccc|c|c|}
    \hline
    \rowcolor[HTML]{B7B7B7}
     Model        &                  & Dataset        &                               & Hyperparameters       & Reasoning \\ \hline
    \rowcolor[HTML]{B7B7B7} 
                  & Subject Count    & Channels       & Train:Test:Validation(\%)     &                       &            \\ \hline
     ResNet50     & \thead {All 10 patients\\ and 4 controls}    & All 10 channels & 80:10:10  & \begin{tabular}[c]{@{}l@{}}Weights:’imagenet’; pooling:’avg’;\\ activation: ‘softmax’;\\optimizer: 'Adam';\\ Loss: ‘categorical\_crossentropy; \\   learning rate: 0.001;\\Patience: 200 epochs; \\ monitoring :’validation accuracy’.\end{tabular}  & \thead{We want to test \\ model performance \\ on multi-dimensional \\ data.i.e. training with \\ all available channels.} \\ \hline

     \thead{ResNet50 \\ ResNet50 \\ResNet50 \\ ResNet50 \\ ResNet50 \\ResNet50 \\ ResNet50 \\ ResNet50 \\ResNet50 \\ ResNet50}   & \thead {All 10 patients\\ and 4 controls}  & \thead{COX4 \\ Dystrophin \\ GRIM19 \\ MTCO1 \\NDUFB8 \\ OSCP \\ SDHA \\ TOM22 \\ UqCRC2 \\ VDAC1  }  & same as above & same as above & \thead{We want to test model \\ performance when trained \\ on individual channels,\\ this will also help with \\ explainability i.e. isolating \\ the relevance of this channel.} \\ \hline

    \end{tabular}
    \newline
    \caption{Experiment Design (Models): model training details for ResNet50.We repeated the above experiments with VGG-16 as well to compare results and decide the models to which we will apply explainability methods.}
    \label{tab:Experiment design model}
\end{table*}

\begin{table*}[]
    \centering
    \begin{tabular}[width=1\textwidth]{|l|l|}
    \hline
    \rowcolor[HTML]{B7B7B7}
      Models Trained on Dataset   & Explainability Methods Applied  \\ \hline
      All patients and control  & \thead{[‘Gradients’, ‘Deconvnet’, ‘Guided Backpropagation’,  ‘input * Gradients’, ‘Deep Taylor’,\\ ‘LRP-Epsilon’, ‘LRP-Z’, ‘LRP-PresetAFlat’, LRP-PresetBFlat’]} \\ \hline

    \end{tabular}
     \newline
    \caption{Experiment Design (Explainable Methods): names of all explainable methods applied on the top 5 (excluding models trained on all channels) performing models.}
    \label{tab:Experiment design explainable methods}
\end{table*}

\section{Data and Methods}
\subsection{Data}
In this study, we use protein expression images from a published dataset \cite{Warren2020DecodingCytometry} that were obtained using IMC. The two main classifications within the dataset are control subject (n=4) and patient (n=10). Patients are further divided into six classes based on their genetic diagnosis (i.e. specific type of mitochondrial disease). We have one SM tissue section for each subject (n=14). For each SM tissue section there are 10 different protein expression images (pseudo images of 10 protein markers generated by IMC). The proteins that were observed are mostly mitochondrial, including subunits of complex I (NDUFB8 and Grim19), complex II (SDHA), complex III (UQCRC2), complex IV (MTCO1 and COX4), complex V (OSCP) and two outer mitochondrial membrane proteins (VDAC1 and TOM22), plus a muscle membrane protein (Dystrophin) as a cell marker, to aid image segmentation. This amounts to a total of 140 images from 14 subjects.

\subsection{Methods}
Selection of DL models for our experiment: As discussed above selection of DL models are usually based on task (our task: image classification), performance (i.e. performance on known benchmark image classification tasks e.g. winners of ILSVRC challenge), amount of training data (for small training data consider transfer learning) and computer resources.

Based on the above considerations we selected VGG and ResNet with transfer learning. While it is usually prescribed to use closely related  transfer learning data but in the absence of any known pre-trained models that were trained on muscle fibres, we selected imagenet weights as initial weights for our models.  We believe the sheer amount of imagenet training data must at least provide some generalisability for our models.

Selection of explainable methods: As discussed above various methods under the three categories tell us different things about the basis of predictions and they compliment each other to further explain the model predictions. But attribute methods give the most information, followed by signal methods and function methods. We want to explore many attribution methods, some signal methods and at least one function method.
Based on above considerations we selected a function method ‘Gradients’ \cite{Smilkov2017SmoothGrad:Noise}, two signal methods ‘Deconvnet’, ‘Guided Backpropagation’ \cite{Zeiler2014LNCSNetworks}\cite{SpringenbergSTRIVINGNET}, and six attribution methods ‘input * Gradients’, ‘Deep Taylor’, ‘LRP-Epsilon’, ‘LRP-Z’, ‘LRP-PresetAFlat’, LRP-PresetBFlat’ \cite{Alber2019INNvestigateNetworks}\cite{Montavon2017ExplainingDecomposition}\cite{Bach2015OnPropagation}. 

\section{Experiments}
Pre-processing: we divided images into smaller sections for ease of training i.e. more augmented data and better convergence. This resulted in 2,730 512x512 image patches distributed across 10 channels.

\subsection{Experiment design  for models} With an aim to classify if an image belongs to ‘patient’ or ‘control’ class, we design following combination of DL model training experiments as detailed in Table \ref{tab:Experiment design model}

\subsection{Experiment design for explainable methods} We restricted application of explainability methods to models trained on single-channel data, as visualisation of input in 10-channels will be complex. We applied all 9 explainability methods on top 5 performing models that were trained on ‘all patients’ dataset as detailed in Table \ref{tab:Experiment design explainable methods}. 

\footnote{Github repository for this paper can be found at \url{https://github.com/atifkhanncl/Explainable_mitoML} }
    
\begin{table*}[]
    \centering
    \begin{tabular}[width=1\textwidth]{|cccrrrrrrr|}
    \hline
    \rowcolor[HTML]{CCCCCC} 
       Model  & Dataset & Channel & \thead{TA \ RS-A(\%)} & \thead{TA \ RS-B(\%)} & \thead{TA \ RS-C(\%)} & \thead{TA \ RS-D(\%)} & \thead{Mean TA(\%)} & \thead{SD TA} & \thead{Var TA}\\ \hline
        VGG16 & All Patients & All Channels & 100 & 98.95 & 98.95 & 98.95 & \cellcolor[HTML]{6AA84F}99.21 & 0.52 & 0.27 \\ \hline
        ResNet50 & All Patients & UqCRC2 & 100 & 92.86 & 100 & 96.43 & \cellcolor[HTML]{6AA84F}97.32 & 3.41 & 11.68 \\ \hline
        ResNet50 & All Patients & GRIM19 & 100 & 92.86 & 92.86 & 96.43 & \cellcolor[HTML]{6AA84F}95.53 & 3.41 & 11.68 \\ \hline
        ResNet50 & All Patients & Dystrophin & 96.43 & 96.43 & 92.86 & 92.86 & \cellcolor[HTML]{6AA84F}94.64 & 2.06 & 4.24 \\ \hline
        ResNet50 & All Patients & OSCP & 92.86 & 96.43 & 92.86 & 92.86 & \cellcolor[HTML]{6AA84F}93.75 & 1.78 & 3.18 \\ \hline
        ResNet50 & All Patients & COX4 & 100 & 96.43 & 85.71 & 89.29 & \cellcolor[HTML]{6AA84F}92.85 & 6.52 & 42.53 \\ \hline
        ResNet50 & All Patients & SDHA & 89.29 & 82.14 & 92.86 & 96.43 & 90.18 & 6.10 & 37.22 \\ \hline
        ResNet50 & All Patients & NDUFB8 & 85.71 & 85.71 & 92.86 & 85.71 & 87.49 & 3.57 & 12.78 \\ \hline
        ResNet50 & All Patients & VDAC1 & 85.71 & 78.57 & 89.29 & 85.71 & 84.82 & 4.49 & 20.20 \\ \hline
        ResNet50 & All Patients & TOM22 & 71.43 & 85.71 & 89.29 & 85.71 & 83.03 & 7.91 & 62.70 \\ \hline
        ResNet50 & All Patients & MTCO1 & 67.86 & 82.14 & 75 & 92.86 & 79.46 & 10.66 & 113.73 \\ \hline
        
    \end{tabular}
    \newline
    \caption{Model Ranking: models trained on ‘AllPatients’ dataset on various channels ordered by mean test accuracy over 4 different training runs which were distinguished by random seeds. * TA, RS, SD and VAR stands for Test Accuracy,  Random Seed, Standard Deviation and Variance respectively. }
    \label{tab:Model ranking AllPatients}
\end{table*}

\begin{figure}[htbp]
\centerline{\includegraphics[width=0.5\textwidth]{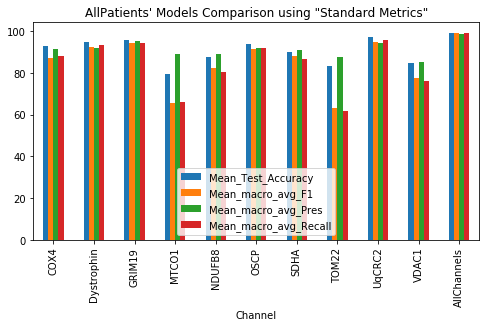}}
\caption{Comparison of models trained on AllPatients dataset using standard metrics.}
\label{Fig:Plot standard metrics allpatients }
\end{figure}

\begin{figure}[htbp]
\centerline{\includegraphics[width=0.5\textwidth]{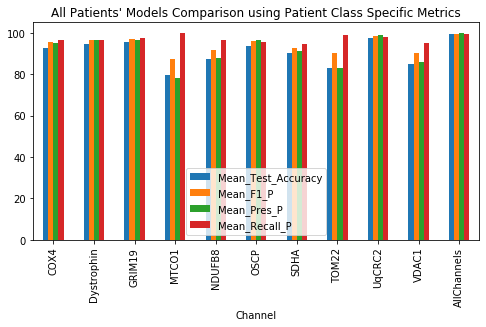}}
\caption{Comparison of models trained on AllPatients dataset using patient class specific metrics.}
\label{Fig:Plot patient metrics allpatients }
\end{figure}

\begin{figure}[htbp]
\centerline{\includegraphics[width=0.5\textwidth]{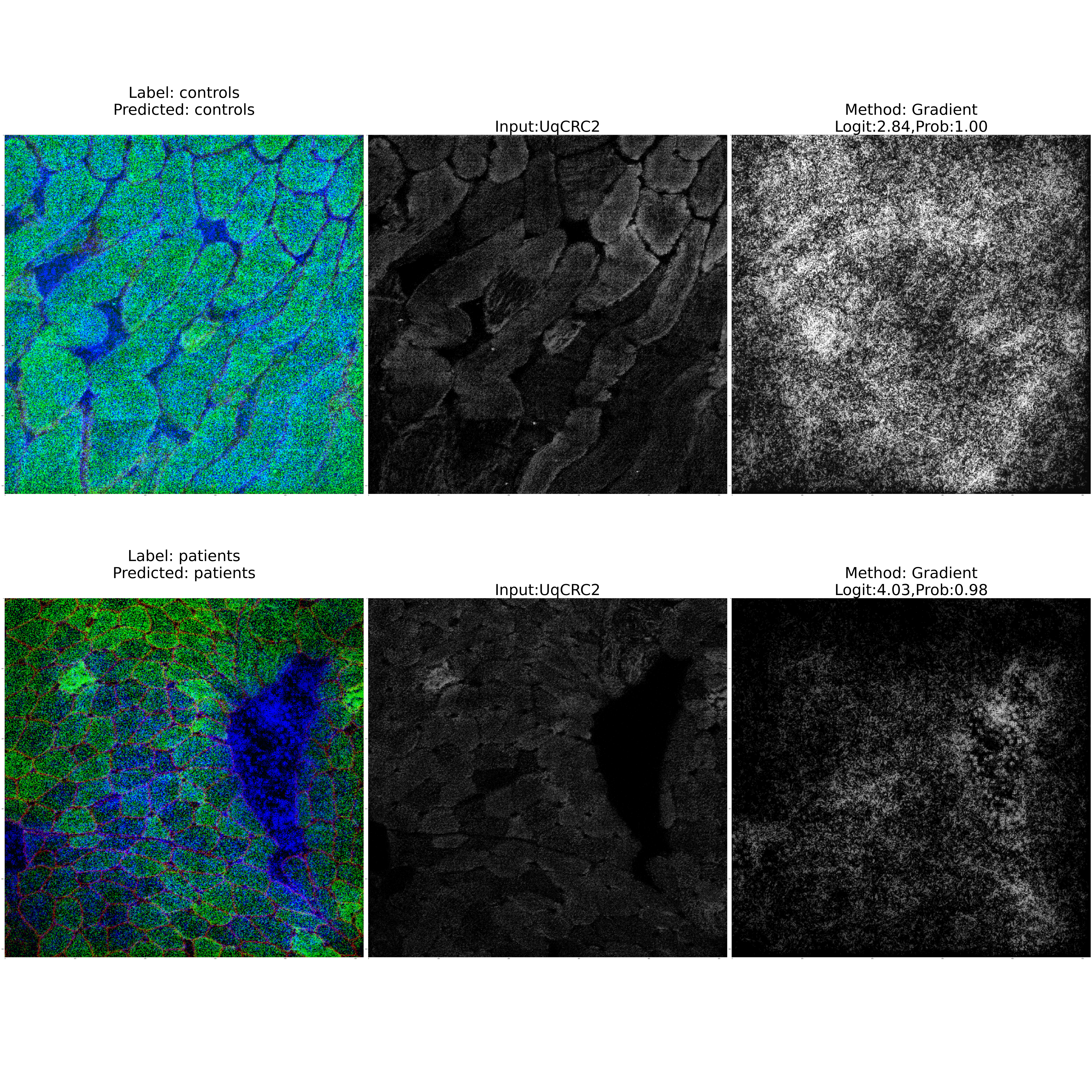}}
\caption{A result from the explainable function ’gradient’ method that shows the input pixels that are most sensitive to the output. Images on the left are constructed by assigning (R,G,B) colours to (fiber membrane, mitochondrial mass, gradient map) for domain experts to visualise the result, images in the middle are input test images and images on the right are gradient map outputs.*model used here was trained on AllPatients dataset and UqCRC2 channel.}
\label{Fig:explainable functionmethod Gradient }
\end{figure}
\begin{figure}[htbp]
\centerline{\includegraphics[width=0.5\textwidth]{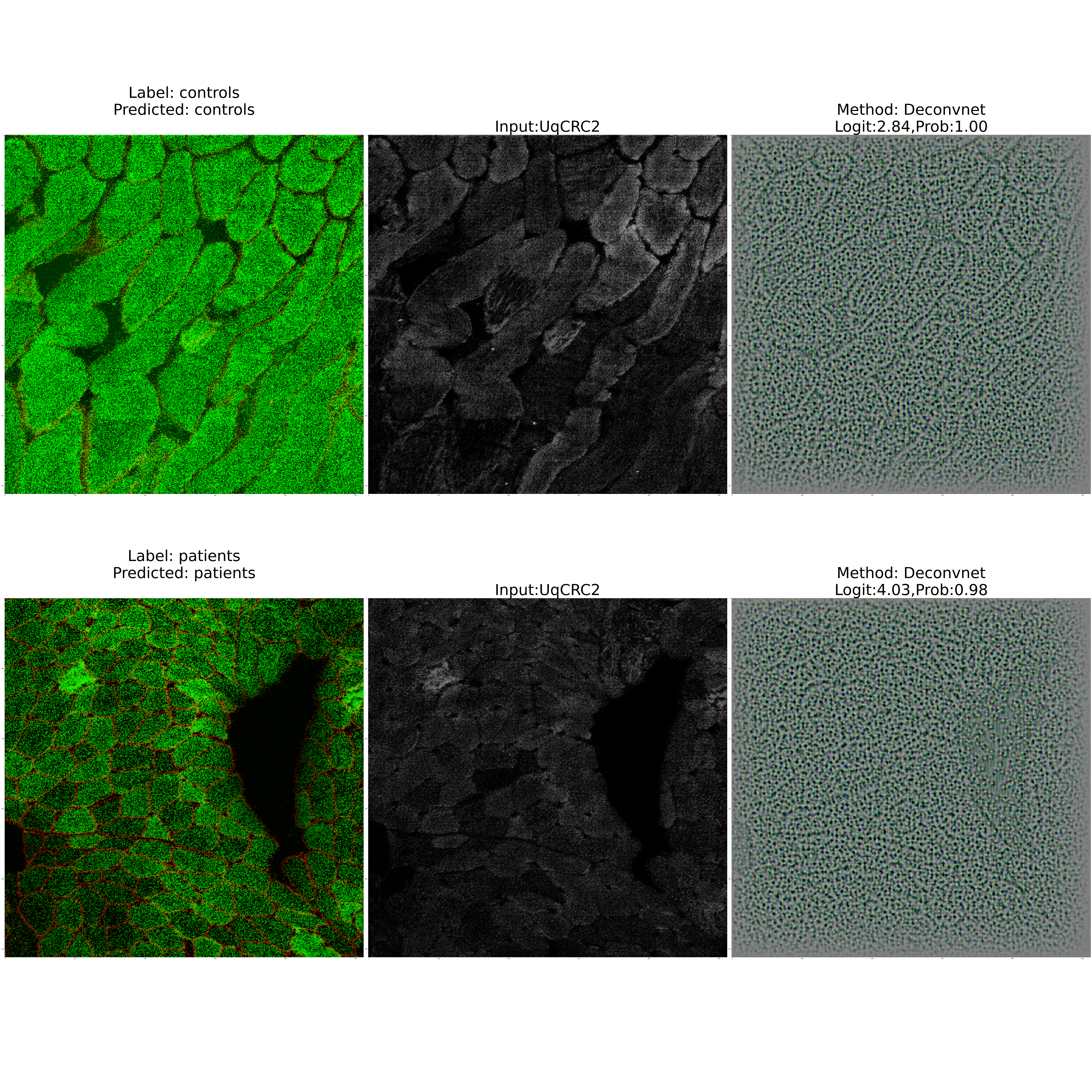}}
\caption{A result from the explainable signal ’Deconvnet’ method that shows the signal/feature/components of the input image that causes the output. Images on the left are constructed by assigning (R,G,B) colours to (Fiber membrane, mitochondrial mass, black) for domain experts to visualise the result, images in the middle are input test images and images on the right are outputs from the explainable method.*We did not combine the explainable map on the left images as this was making visualisation difficult, model used here was trained on AllPatients dataset and UqCRC2 channel.}
\label{Fig:explainable signal method Deconvnet }
\end{figure}

 \begin{figure}[htbp]
\centerline{\includegraphics[width=0.5\textwidth]{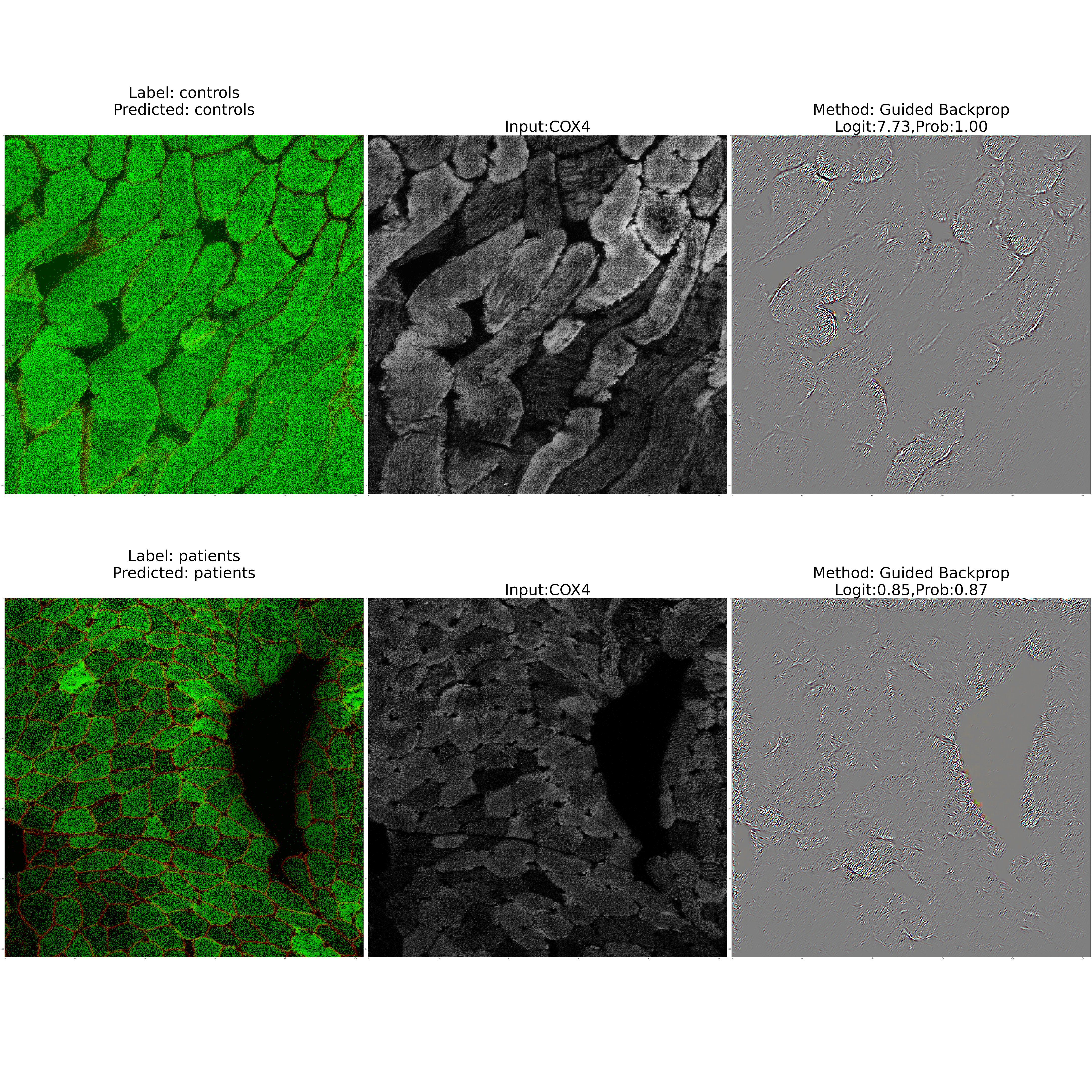}}
\caption{A result from the explainable signal ’Guided Backpropagation’ method that shows the signal/feature/components of the input image that causes the output. Images on the left are constructed by assigning (R,G,B) colours to (Fiber membrane, mitochondrial mass, black) for domain experts to visualise the result, images in the middle are input test images and images on the right are outputs from the explainable method.*We did not combine the explainable map on the left images as this was making visualisation difficult, model used here was trained on AllPatients dataset and COX4 channel.}
\label{Fig:explainable signal method Guided backpropogation }
\end{figure}

\section{Results and Discussion}

\subsection{Classification results}
The classification accuracy we observed in the 'AllPatients' dataset was better than expected. Achieving a mean test accuracy of $>$99\% when trained on all channels and we also observed high mean test accuracy even when trained on some individual channels e.g. 97.3\% using UqCRC2 channel.

\begin{figure}[htbp]
\centerline{\includegraphics[width=0.5\textwidth]{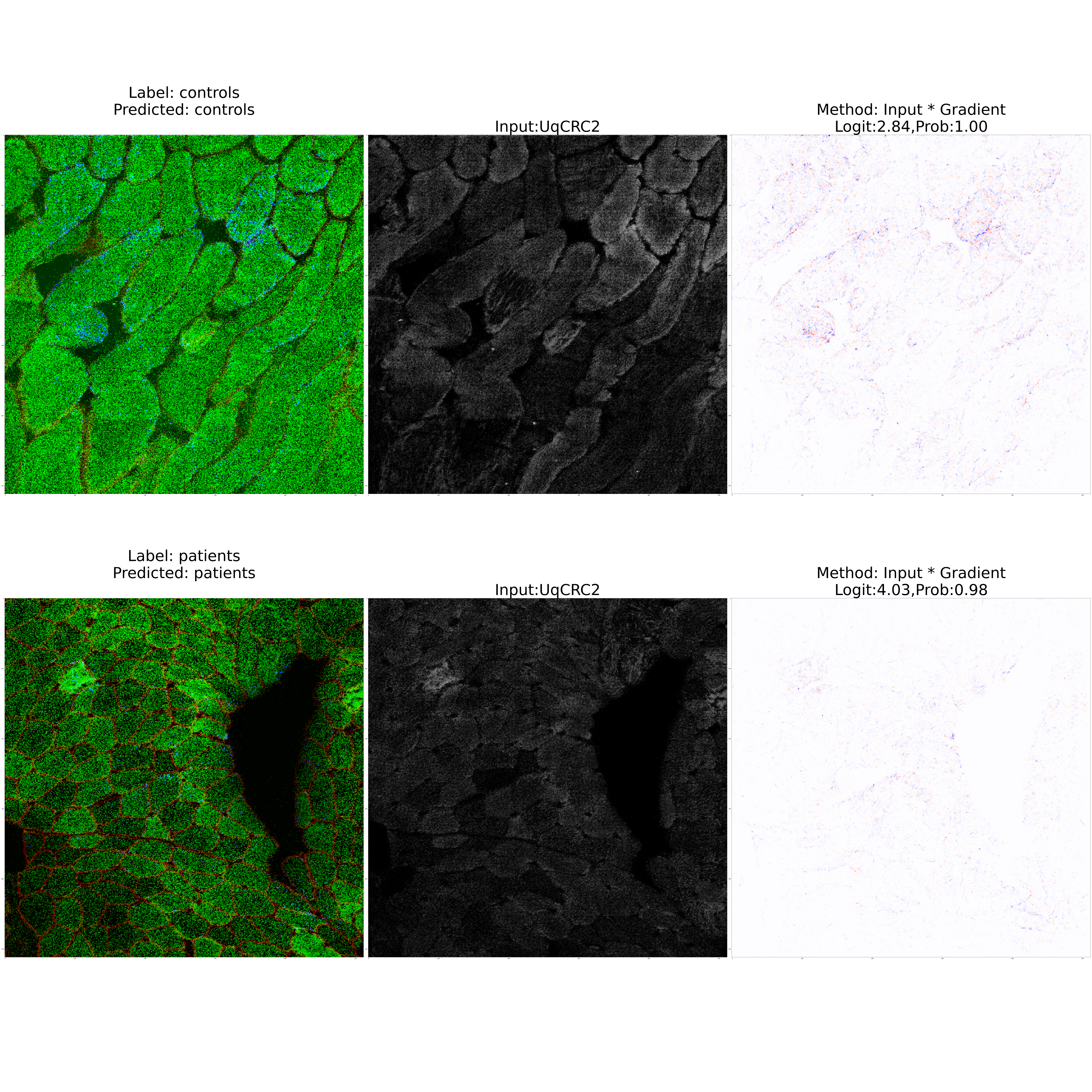}}
\caption{A result from the explainable attribution ’Input * Gradient’ method that detects both the features in the input and their importance/contribution towards the output. Images on the left are constructed by assigning (R,G,B) colours to (fiber membrane, mitochondrial mass, feature importance map) for domain experts to visualise the result, images in the middle are input test images and images on the right are outputs from the  ’Input * Gradient’ method. *model used here was trained on AllPatients dataset and UqCRC2 channel.}
\label{Fig:explainable attribution method inputGradient }
\end{figure}
Attribution methods: These methods consider many signals/features in the model's network and associate these features with location (pixels) to build the explainability map (EM). As these methods combine many features and associated pixel importance the resulting EM is distilled to most important feature-linked pixels. This can be both good or bad, as it might either make the EM uncover the model’s basis precisely or it might be so abstract that it hides the required information to uncover the model's basis. The results from all 6 attribution methods applied to models did show these EMs to be distilled features linked pixel maps i.e. as opposed to other two methods the visualisations are much cleaner and pointing to fewer pixels, as seen in one of the attribution methods presented in Figure \ref{Fig:explainable attribution method inputGradient }. 
     
\begin{figure}[htbp]
\centerline{\includegraphics[width=0.5\textwidth]{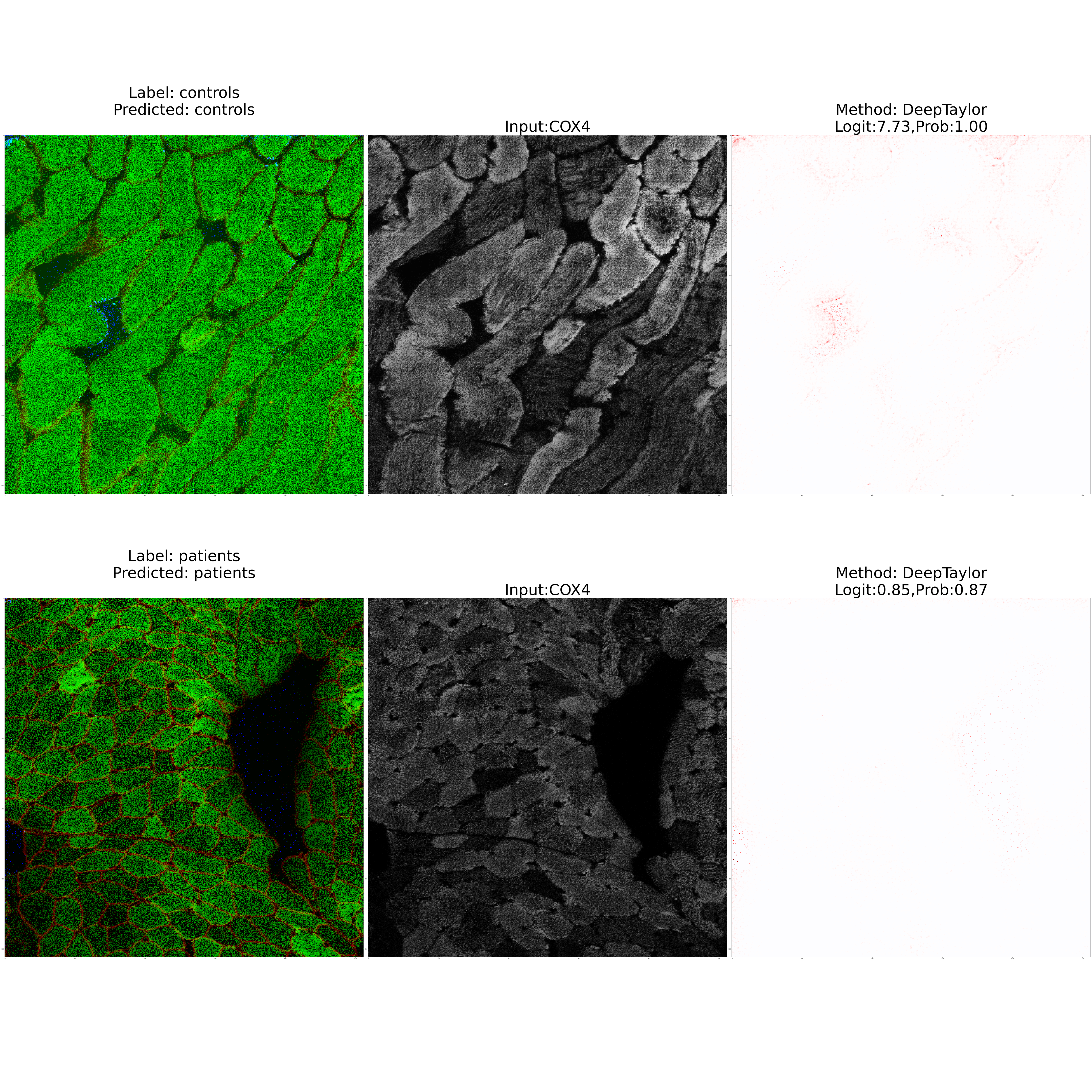}}
\caption{A result from the explainable attribution ’Deep Taylor’ method that detects both the features in the input and their importance/contribution towards the output. Images on the left are constructed by assigning (R,G,B) colours to (fiber membrane, mitochondrial mass, feature importance map) for domain experts to visualise the result, images in the middle are input test images and images on the right are outputs from the  ’Deep Taylor’ method.*model used here was trained on AllPatients dataset and COX4 channel.}
\label{Fig:explainable attribution method DeepTalor }
\end{figure}

\begin{figure}[htbp]
\centerline{\includegraphics[width=0.5\textwidth]{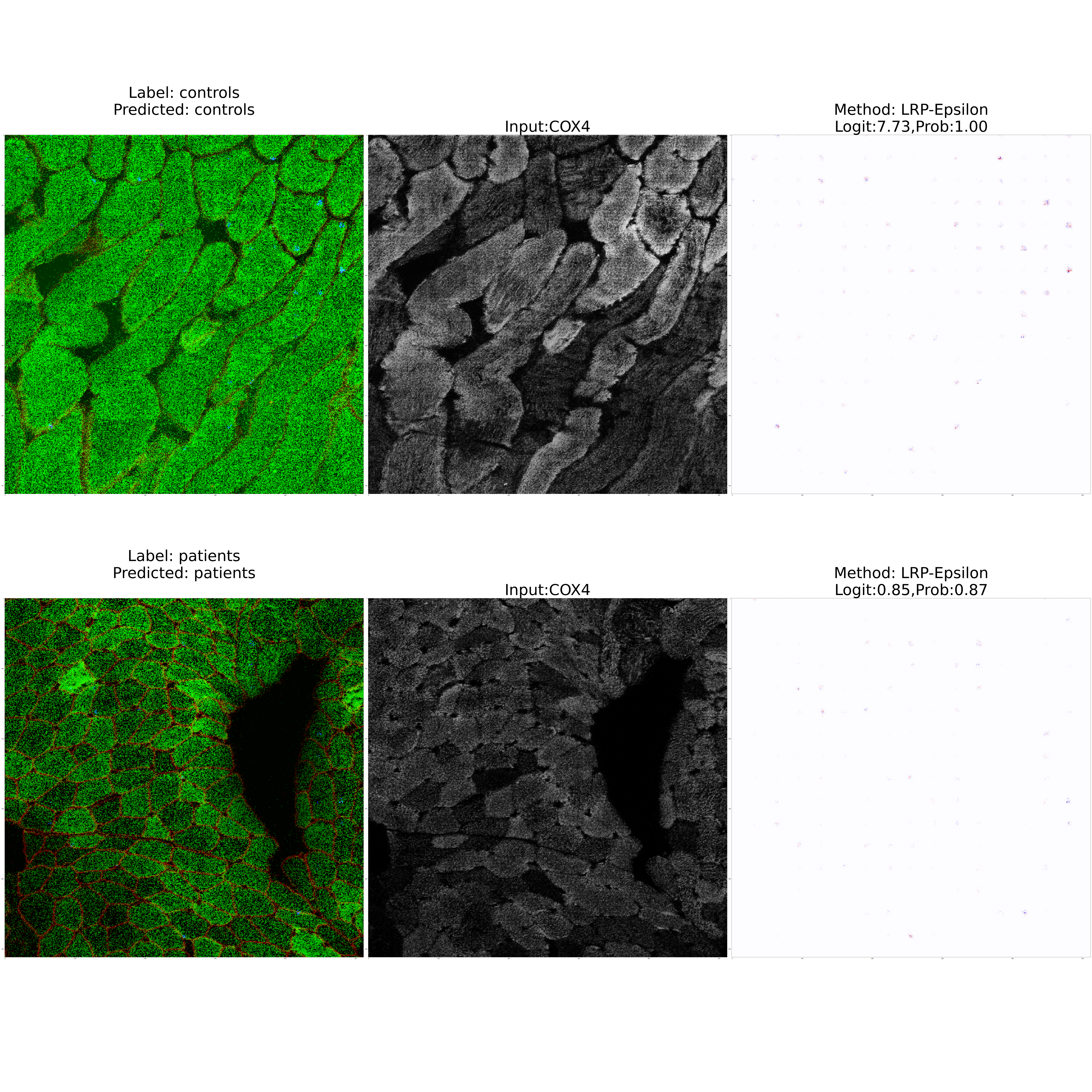}}
\caption{A result from the explainable attribution ’LRP-Epsilon’ method that detects both the features in the input and their importance/contribution towards the output. Images on the left are constructed by assigning (R,G,B) colours to (fiber membrane, mitochondrial mass, feature importance map) for domain experts to visualise the result, images in the middle are input test images and images on the right are outputs from the  ’LRP-Epsilon’ method.*model used here was trained on AllPatients dataset and COX4 channel.}
\label{Fig:explainable attribution method LRP-Epsilon }
\end{figure}

\begin{figure}[htbp]
\centerline{\includegraphics[width=0.5\textwidth]{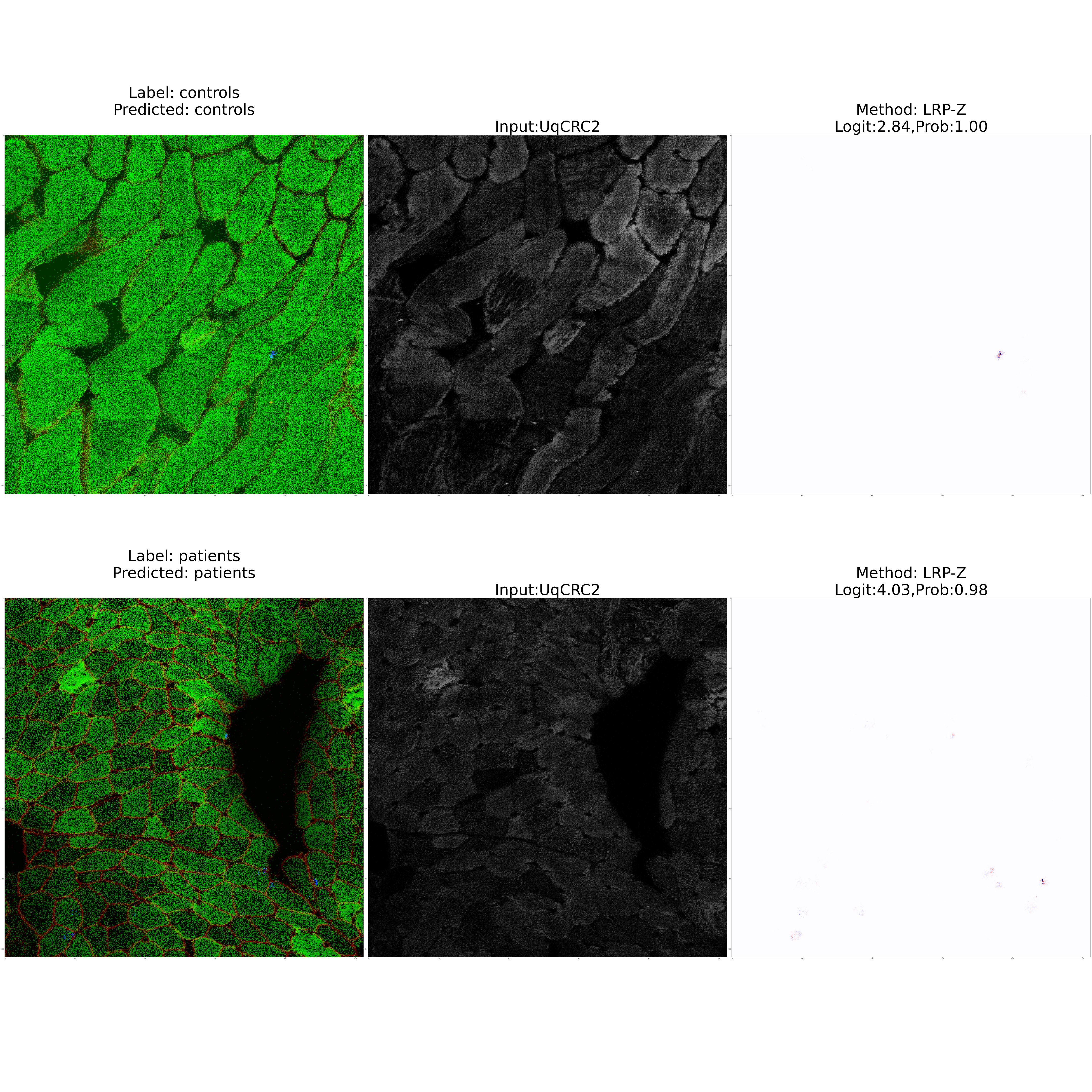}}
\caption{A result from the explainable attribution ’LRP-Z’ method that detects both the features in the input and their importance/contribution towards the output. Images on the left are constructed by assigning (R,G,B) colours to (fiber membrane, mitochondrial mass, feature importance map) for domain experts to visualise the result, images in the middle are input test images and images on the right are outputs from the  ’LRP-Z’ method.*model used here was trained on AllPatients dataset and UqCRC2 channel.}
\label{Fig:explainable attribution method LRP-Z }
\end{figure}
\begin{figure}[htbp]
\centerline{\includegraphics[width=0.5\textwidth]{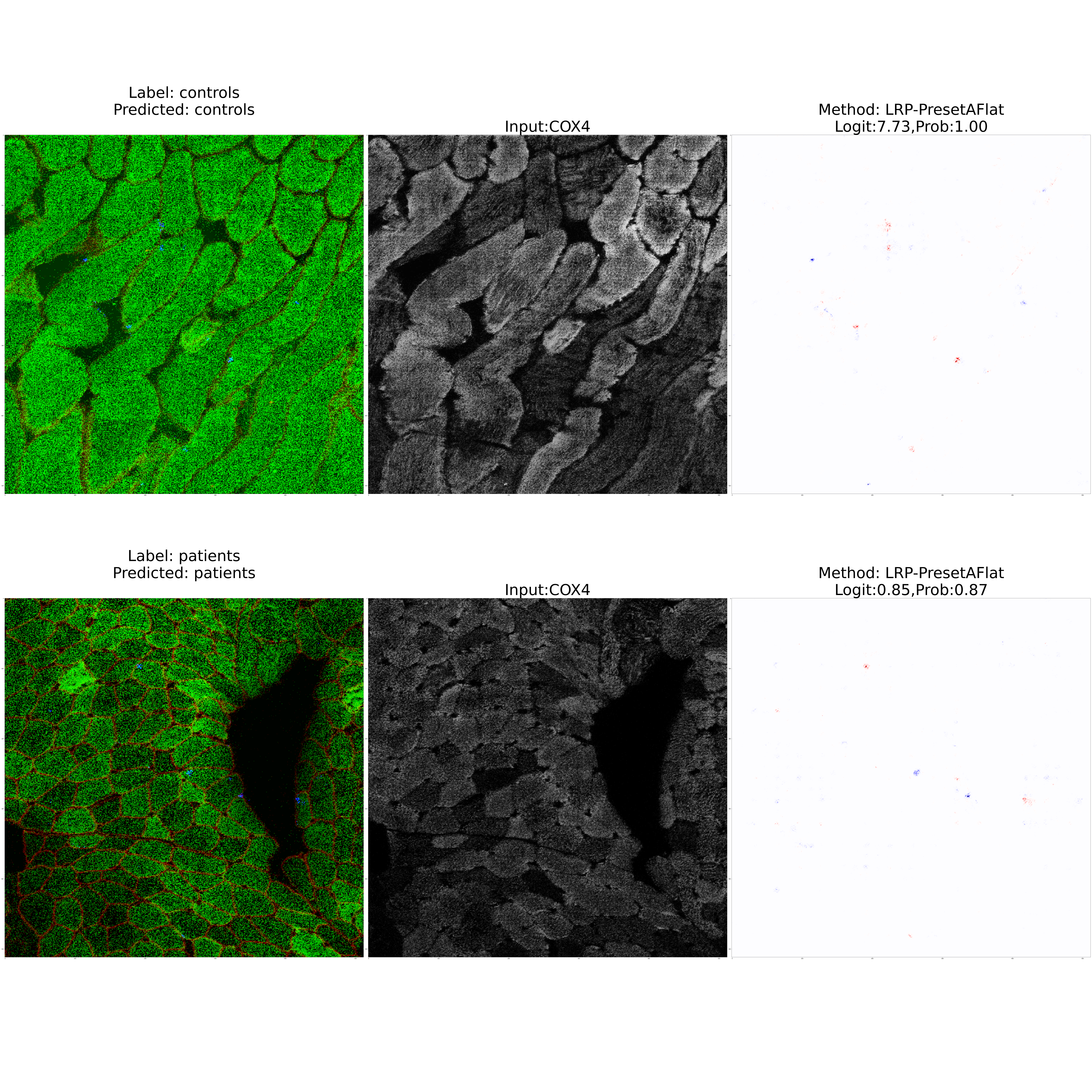}}
\caption{A result from the explainable attribution ’LRP-PresetAFlat’ method that detects both the features in the input and their importance/contribution towards the output. Images on the left are constructed by assigning (R,G,B) colours to (fiber membrane, mitochondrial mass, feature importance map) for domain experts to visualise the result, images in the middle are input test images and images on the right are outputs from the  ’LRP-PresetAFlat’ method.*model used here was trained on AllPatients dataset and COX4 channel.}
\label{Fig:explainable attribution method LRP-PresetAFlat }
\end{figure}
\begin{figure}[htbp]
\centerline{\includegraphics[width=0.5\textwidth]{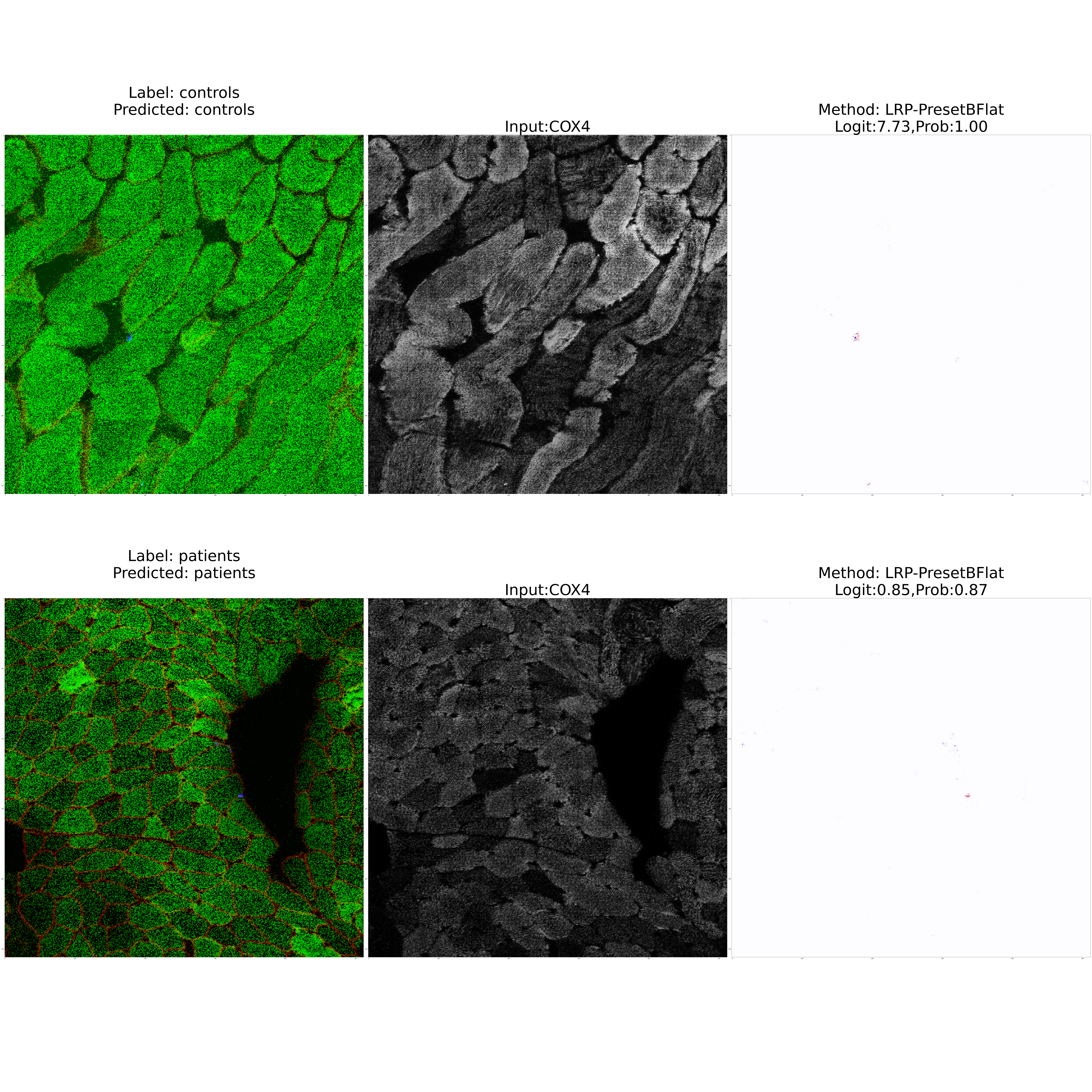}}
\caption{A result from the explainable attribution ’LRP-PresetBFlat’ method that detects both the features in the input and their importance/contribution towards the output. Images on the left are constructed by assigning (R,G,B) colours to (fiber membrane, mitochondrial mass, feature importance map) for domain experts to visualise the result, images in the middle are input test images and images on the right are outputs from the  ’LRP-PresetBFlat’ method.*model used here was trained on AllPatients dataset and COX4 channel.}
\label{Fig:explainable attribution method LRP-PresetBFlat }
\end{figure}

As seen in Table \ref{tab:Model ranking AllPatients}. The models trained ‘AllPatients’  dataset with all channels achieved near perfect mean test accuracy (99.2\%) with low standard deviation (0.525). This ranking is expected as the model has access to all channels and so all available information, but the level of accuracy is impressive and better than expected.

Agreement between ‘standardised’ accuracy evaluation metrics: As seen in Figure \ref{Fig:Plot standard metrics allpatients } there is, in general, agreement between other evaluation metrics such as ‘macro average of F1’, ‘macro average of precision’ and ‘macro average of recall’ with the test accuracy. This is reassuring to us with regards to performance of models across both the classes.

Comparing patient class specific evaluation metrics with test accuracy: As seen in Figure \ref{Fig:Plot patient metrics allpatients } for models trained on ‘Allpatients’ dataset the patient class specific metrics such as ‘Precision’, ‘Recall’ and ‘F1’ scores of patient class are broadly in agreement with respective test accuracy scores. Biologically it is important for a classifier to perform better on patient class than control due to heterogeneity of mitochondrial disease \cite{Warren2020DecodingCytometry} and a similar or higher accuracy on patient class when compare to class adjusted metrics such as test accuracy is a positive result.
 
\subsection{Explainable methods}
All explainable methods worked as intended i.e. providing various ways to visualise the model’s basis for prediction. 
Function method: As seen in Figure \ref{Fig:explainable functionmethod Gradient } the output from gradient functional method provides visualisation of pixels in two input test images (a control and patient) that influence the output classifier the most. We applied this method on top 5 models from both datasets (excluding models’ trained on all channels) and observed the gradient map (GM) to be varied e.g. when comparing the GM as seen Figure \ref{Fig:explainable functionmethod Gradient } across models trained on different channels we saw shift of model focus to different patterns i.e. in some cases inter-fiber areas are highlighted in others it highlights near uniformly random pixels over the image. This might make sense as all these models are trained on different channels and might be basing the prediction on different features. But with our raw dataset that does not have any identifiable segmentation, it is hard to associate the model’s pixel GM to the object (fibres) in the image to make sense of it.

Signal methods: These methods try to distil a signal or feature in the model’s network and associated importance of pixels within that feature. In all our experiments, including one presented in Figure \ref{Fig:explainable signal method Deconvnet } we observed both signal methods but especially the Deconvnet picking the morphology/shape of fibres as the signal which is reassuring and biologically makes sense. But the importance of pixels within these features is hard to comprehend as pixel importance (signified by colour and intensity) seems to be distributed all over the image.

\subsection{Utility of these techniques as a tool to analyse high dimensional IMC data}
Explainability methods that try to understand the basis for predictions from the ResNet50 model applied to IMC image stacks from patient and control subjects lead to a diverse and not entirely consistent set of conclusions.  Several of the methods we investigated highlight biological features of the imaged muscle tissue, which is reassuring.  However, some methods highlight different image features and some seem to ignore biologically relevant signal altogether.  This lack of consistency between the methods makes it quite challenging to understand the utility of each explainability method.

For example, the Gradient method in Figure \ref{Fig:explainable functionmethod Gradient } and the Deep Taylor method in Figure \ref{Fig:explainable attribution method DeepTalor } highlight gaps in tissue.  This is the opposite of what we might have expected, since these biological experiments are set up to observe protein signal within cells within tissue, not holes in tissue.  It could be that the prevalence of gaps or holes in tissue sections is different in patients compared to control.  However, if that is the case, it seems likely that the difference would be a result of experimental artefacts: slight differences in the ways that the data are collected, the exact muscle used or the process by which the tissue is collected. As such we would not expect that these predictions are robust to different lab experiments: i.e. would likely not have the same predictive power with different patients, or in experiments carried out by different labs, or even repeated in the same patients in the same lab.

The Deconvnet method in Figure \ref{Fig:explainable signal method Deconvnet } seems to focus on the average pixel intensity across images with slight downweighting of regions where tissue is missing.  

Guided Backpropagation highlights cell membranes in controls Figure \ref{Fig:explainable signal method Guided backpropogation }. Cell membranes are relevant in this context as mitochondria are more abundant in the region just inside the cell membrane. This phenotype is further enhanced in ragged red fibres (RRFs) which are a pathological hallmark in some patients and are cells with an increase in mitochondria most prominently around the edges of the cell. We expect that RRFs should be randomly distributed around the tissue in patients (where OXPHOS defects are more prevalent).  However, spatial distribution of cells where the region adjacent to the membrane is highlighted by Guided Backpropagation is not random: cells which are adjacent to gaps in the tissue are always selected, this may suggest that a contributing factor in the selection of areas by this method is contrast between the signal either side of the membrane. Again, methods that focus on gaps in tissue are likely to be responding to experimental artefacts and predictions are unlikely to be robust.

The Input * Gradient method in Figure \ref{Fig:explainable attribution method inputGradient } seems to focus on average fibre intensity in controls vs fibre membranes in patients.  Unlike the Guided Backpropagation this method does not seem to focus more on the cells adjacent to gaps in the tissue. Furthermore, the areas selected seem to focus on specific regions adjacent to the membranes in some cells, which may fit with some hypotheses for how OXPHOS deficiency develops \cite{vincent2018subcellular}.
LRP-Z seems to focus on the mean image-level pixel intensity in patient images, while focussing on some points in control images, however these points seem surprisingly sparse and also seem randomly distributed as seen in Figure \ref{Fig:explainable attribution method LRP-Z }, which does not seem relevant.

LRP-Epsilon in Figure \ref{Fig:explainable attribution method LRP-Epsilon } focuses on an artificial regular grid of points in the images, which is not possible to understand from a biological perspective.

LRP-Preset AFlat in Figure \ref{Fig:explainable attribution method LRP-PresetAFlat } and LRP-Preset BFlat in Figure \ref{Fig:explainable attribution method LRP-PresetBFlat } seem to highlight small regions of cell perimeter of patient images which could correspond to myonuclei.  This is an intriguing observation, and again may be consistent with the perinuclear niche hypotheses of Vincent et al.\cite{vincent2018subcellular}. This method works well for COX4 in particular, which is not one of the proteins we would expect to be most useful in distinguishing patients and controls.

Overall the two LRP-Preset methods seem the most promising and most consistent with our incomplete understanding of how OXPHOS defects arise in these patient cells. Generally, however, the contradictions between these methods is a cause for concern from a biomedical perspective.  It is perhaps the case that this kind of scientific research, which has to be exploratory to a large degree, is not an ideal use case for explainable Deep Learning methods, since our incomplete understanding of mechanisms by which the patterns in these images arise means that we do not have a firm gold standard or ground truth from which to assess the relative merit of different methods.  It is likely that, to fully explore the utility of explainability maps, it will be important to make visual and statistical comparisons between them and the output from established approaches based on fibre segmentation and statistical analysis.

\section{Conclusion and Future Work}

In this paper we have seen that DL models especially when trained on all available channels perform with high accuracy attaining mean test accuracy of up to 99.2\%. This is impressive but the main objective of analysing IMC images is to find the relationship between different protein markers’ expression levels to mitochondrial diseases. Highly accurate DL models are useful for prediction, but not for improving our understanding of mitochondrial disease as these are black boxes that do not provide the basis for their predictions. To solve this we applied various state of the art explainable methods to these models.      
        	     
The explainable methods we applied provided various visualisation maps to understand the basis of predictions of the models. While these maps give some information about features and location of importance in a given IMC image, this is very limited in context i.e. it needs refining both in terms of details in visualisation it provides but also extending this to interrogate models trained on all 10 or even more channels. Unlike  other use cases of explainable methods where they are used to validate a model's basis for prediction  by comparing the ’logic’ used by the model to that of humans, the problem we are tackling is more difficult, we have only partial ground truth human logic to compare against. We believe the work accomplished in this paper is introductory upon which a combination of techniques can be developed along with prescription of their use to uncover and validate the ‘logic’ returned by the techniques.

For future work we want to conduct experiments with segmented SM data that will allow us to restrict the feature space to the SM level rather than the whole tissue or clusters of SMs. This we believe will result in some attribution based explainable methods to be more informative. With regards to extending the application of explainable methods to multichannel models we want to experiment with various combinations of RGB images made by combining various protein expression channels as inputs. This can be further extended to include more channels but aggregating visualisation maps from models trained on various channels. With regards to unavailability of ground truth human logic for validation, we propose comparing with the pattern/logic identified at SM level by current statistical approaches.

\section{Acknowledgements}
This work was supported by the Wellcome Centre for Mitochondrial Research (203105/Z/16/Z), EPSRC Centre for Doctoral Training in Cloud Computing for Big Data. AEV is in receipt of a Sir Henry Wellcome Fellowship (215888/Z/19/Z).
\bibliographystyle{ieeetr}
\bibliography{references_use}

\begin{thebibliography}{10}
\bibitem{AnTechnology}
``{An artificial intelligence tool that can help detect melanoma | MIT News |
  Massachusetts Institute of Technology}.''
\end{thebibliography}

\end{document}